\documentclass[12pt]{extarticle}
\usepackage{amsfonts}
\usepackage{natbib}
\usepackage[margin=1in]{geometry}
\usepackage{url}
\usepackage{amsmath}
\usepackage{amsthm}
\usepackage{graphicx}
\usepackage{float}
\usepackage{booktabs}

\title{Probabilistic learning of boolean functions applied to the binary classification problem with categorical explanatories}

\author{P. Hubert$^1$}
\date{%
    $^1$ Research coordinator at \textit{Instituto Paulo Gontijo}. Professor at the Technology and Data Science Department, FGV/EAESP}

\begin{document}

\maketitle

\section{Introduction}\label{sec:intro}

Consider a sample $y \in \{0,1\}^n$ generated by two different Bernoulli distributions with parameters $\pi_0$ and $\pi_1$, and consider the set $S \subset \{1,...,n\}$ as the set of all indices $i$ such that $P(y_i) = \pi_1$. Assuming that the components of the vector $y_i$ are conditionally independent given $\theta = (S, \pi_0, \pi_1)$, the likelihood function is the product of two Binomial distribution functions, and will attain a global maximum at the set $S = L(y) = \{i: 1 \leq i \leq n \,\wedge \, y_i = 1\}$ (let's call this set the \emph{on-set} of the vector $y$), with maximum likelihood estimators given by $\hat{\pi}_0 = 0$ and $\hat{\pi}_1 = 1$.

Now consider a design matrix $X \in \mathbb{R}^{n\times p}$ and a function $f:\mathbb{R}^p\to \{0,1\}$ such that $\psi(X_i) = 1 \iff i \in S$, where $X_i$ is the $i$-th row of $X$. Again, if the function $f$ is not constrained in any way, the problem is the same and the same trivial solution applies, with function $f$ defined only in the set of rows of $X$. In this extreme case, the solution will usually not generalize well, and also will not provide any interesting interpretation (since $f$ is just an enumeration based on the on-set of $y$). 

Standard methods for the binary classification problem are concerned with the task of estimating $f$ constraining it in different ways such that this trivial solution (associated with the problem of overfitting) is avoided. This constraints can take many forms; one of the most studied and applied is the use of a linear model with the vector of coefficients constrained to lie in the interior of a ball with a given radius (the LASSO and Ridge regression models). Analysis then proceeds by finding $f$ inside the feasible set such that the distance $\lvert \hat{\pi}_1-\hat{\pi}_0\rvert$ between the MLE for $\pi_0$ and $\pi_1$ is maximized.

In what follows we analyze the case where $X$ is itself a binary matrix, that is $X \in \{0,1\}^{n\times p}$. In this case, $f$ is a boolean function, lying in a discrete and finite space with $2^p$ points. The complete model can be thought of as a table with $2^p$ rows, each row containing one combination of the values of the $p$ binary explanatory variables, plus one column that takes either the value $\pi_0$ or $\pi_1$, representing the probability of the Bernoulli associated with that combination of values. The binary function $f$ assigns to each $x \in \{0,1\}^p$ the value $1$ if the corresponding probability is $\pi_1$.

One of the main appeals of this setup is the well known fact that any boolean function $f$ admits at least one representation as both a disjunction of conjunctions (its \emph{disjunctive normal form}, DNF)

\begin{equation}
\psi(x) = \bigvee_{i=1}^k \left( \bigwedge_{j \in S_i}x_j\bigwedge_{l \in \bar{S}_i}\bar{x}_j\right)
\end{equation}

and a conjunction of disjunctions (its \emph{conjunctive normal form}, CNF)

\begin{equation}
\psi(x) = \bigwedge_{i=1}^k \left( \bigvee_{j \in S_i}x_j\bigvee_{l \in \bar{S}_i}\bar{x}_j\right)
\end{equation}

The conjunction terms in a DNF are called \emph{implicants}, because $\bigwedge_{j \in S_i}x_j\bigwedge_{l \in \bar{S}_i}\bar{x}_j = 1 \implies \psi(x) = 1$. An implicant $S = x_ix_j...\bar{x}_l$ is said to be \emph{prime} if no conjunction obtained by eliminating one literal from $S$ is itself an implicant. A DNF formed by prime implicants is called a prime DNF.

Now given design matrix $X$ and response vector $y$ and assuming that no two rows of $X$ are the same, one can easily prove that there will always be a boolean function $f$ that is a perfect discriminator, i.e., $\psi(x_i) = y_i$; to see this, consider a binary vector $b \in \{0,1\}^n$, and define the on-set $D_1(b)$ of $b$ to be the set $D_1(b) = \{i \in n : b_i = 1\}$, that is, the set of $b$ positions that take the value $1$. Similarly, define the off-set $D_0(b)$ by the set of $b$ positions that take the value $0$. Then a perfect discriminator can be written as

\begin{equation}
f(x) = \bigvee_{x \in T(X,y)} \left( \bigwedge_{j \in D_1(x)}x_j\bigwedge_{l \in D_0(x)}\bar{x}_j\right)
\end{equation}

where $T(X,y) = \{X_i \,:\, i \in D_1(y)\}$ is the set of all rows of $X$ corresponding to positive values in $y$.

What this function does is to form a disjunction between all $X$ rows corresponding to positive values of $y$; it is actually equivalent to a simple enumeration of all positive individuals (positive individual is any $X_i$ such that $y_i = 1$). Of course this function will not generalize well, since it will assign the value $0$ for whatever vector that is not a positive row of $X$.

As discussed above, an usual strategy to avoid overfitting and enhance generalization power is to constrain the function $f$. Following \cite{Muselli2005}, the first restriction we adopt is to \emph{monotone} boolean functions. Monotone boolean functions have the attractive property that their normal forms representations do not depend on the negation operation; that is, any boolean function of the form

\begin{equation}\label{eq:mbf}
y = \psi(x) = \bigvee_{i=1}^k \bigwedge_{j \in S_i}x_j
\end{equation}

is monotone, and any monotone boolean function has a representation in the form \ref{eq:mbf}. 

This however is not a decisive restriction, since it is always possible to augment the dataset by including the negation of each original variable as a new variable, and any boolean function on the original set can be uniquely mapped to a monotone boolean function on the augmented set. Therefore, it is advisable to restrain the class of boolean functions even more.

In the spirit of the LASSO and Ridge regression models, one can define some norm on the space of (monotone) boolean functions, and constrain $\psi$ to lie inside a ball of a fixed radius, according to this norm. Two obvious norms are the number of terms in the DNF of $\psi$, and the total number of atomic variables involved in this same normal form. 

This paper aims to offer an analysis of this problem taking a probabilistic approach. This means that we incoporate the norms indicated above into kernels of prior distributions, and propose posterior inference over the set of monotone boolean functions. 

The biggest challenge in the estimation of these models is the combinatorial nature of all optimization procedures involved. Direct methods suffer from non-polynomial computational complexity; the adoption of a probabilistic setup and simulation algorithms such as MCMC (with polynomial-bounded complexity, see for example \cite{Roberts2016}) or stochastic search methods are useful resources in these kinds of problems.

\subsection{Applications}

This model can be applied to the analysis of any binary classification problem with categorical explanatory variables (by the use of a proper encoding to transform each $m$-category variable into $m-1$ binary variables). 

Our main motivation, however, comes from the problem of estimating \emph{polygenic markers} from genomics data. This is a topic that is currently the subject of great research interest (see for instance recent works by \cite{Mavaddat2019, Shang2019, Yin2019,Haws2015}).

In this context, the goal is to find combinations of genetic mutations (usually single nucleotid polymorphisms - SNP) that are jointly associated with a given phenotype (for instance the presence of a disease), while each single mutation taken by itself has little or no marginal effect on the same outcome.

One the most common approaches is to build a \emph{polygenic risk score}: a predictive estimation of the probability of the phenotype given the genotype. The polygenic part of the score indicates that the estimation model includes interaction terms.

More than the derivation of a polygenic risk score, however, our analysis aims at highlighting specific regions of the genome that together form a polygenic marker. There is to say: besides predictive power, the goal in this work is to allow for direct interpretation of the model's results in substantive terms, such that biologists and other specialists are able to directly evaluate the biological likelihood of the markers obtained. In this sense, an easily interpretable model is much more useful for the development of treatments than a strictly predictive model with many terms, non-linearities, etc.

This is the reason why we have chosen to adopt the discrete formulation of the problem in terms of boolean function estimation. As discussed in section \ref{sec:intro}, the DNF form of a boolean function is immediately interpretable as a logic sentence whose terms can be taken as polygenic markers (because they require that the individual is mutated in position $i$ \textbf{and} $j$ \textbf{and} $k$ etc. of the genome).

\subsection{Structure of the paper}

Section \ref{sec:model} formalizes the probabilistic model. Section \ref{sec:post} introduces the methods for posterior inference for this probabilistic model; section \ref{sec:results} presents experimental results with simulated data, and using UCI's Mushroom dataset. Section \ref{sec:conclusion} concludes the paper.

\section{Probabilistic model}\label{sec:model}

Consider a data set $X \in \{0,1\}^{n\times p}$, $y \in \{0,1\}^n$ consisting of $p$ binary features measured on $n$ individuals, and a single binary outcome $y_i$ for each individual. Given a boolean function $\psi:\{0,1\}^p\to\{0,1\}$, and two parameters $0 \leq \pi_0 \leq \pi_1 \leq 1$, the $y_i$ are conditionally independent Bernoulli variables with probability function given by 

\[
P(y = 1 \mid x, \pi_0, \pi_1, \psi) = 
\begin{cases}
   \pi_0,& \text{if } \psi(x) = 0\\
   \pi_1,& \text{if } \psi(x) = 1
\end{cases}
\]

If $\psi$ is written in its DNF, and assuming it is monotone, then it is completely defined by a set $\omega = \{S_1, S_2, ..., S_m\}$, $S_i \subset \{1,...,p\}$, where $m$ is the number of conjunction terms, and each $S_i$ is the subset of variables involved in each implicant of (the DNF of) $\psi$. From now on, we will use $\omega$ and $\psi$ as equivalent formulations of the same mathematical object. We will also call samples $i$ with $y_i = 1\,(0)$ positive (negative) samples, and samples $i$ with $\psi(x_i)=1\,(0)$ marked (unmarked) samples.

\subsection{Likelihood}
The likelihood for this model, given data $X \in \{0,1\}^{n\times p}, y\in\{0,1\}^n$ and assuming conditional independence of the $y_i$, is

\begin{equation}\label{eq:like}
P(y \mid X, \psi, \pi_0, \pi_1) = \prod_{i=1}^n \left[\pi_0^{(1-\psi(x_i))}\pi_1^{\psi(x_i)}  \right]^{y_i}\left[ (1-\pi_0)^{(1-\psi(x_i))}(1-\pi_1)^{\psi(x_1)}   \right]^{(1-y_i)}
\end{equation}

That is, the likelihood is the product of two Bernoulli likelihoods, one for marked individuals and another for unmarked ones.

After some rearrangement, equation \ref{eq:like} can be written as

\begin{equation}\label{eq:like2}
P(y \mid X, \psi, \pi_0, \pi_1) = \left(1-\pi_0\right)^n\prod_{i=1}^n\left[ \left(\frac{\pi_1(1-\pi_0)}{\pi_0(1-\pi_1)}\right)^{\psi_iy_i} \left(\frac{1-\pi_1}{1-\pi_0}\right)^{\psi_i} \left(\frac{\pi_0}{1-\pi_0}\right)^{y_i} \right]
\end{equation}

The log-likelihood takes the form

\begin{align}
\begin{split}
logP(y\mid X, \theta) = &n\cdot log\left(1-\pi_0\right) + \\
& log\,\frac{o_1}{o_0}\sum_{i=1}^n\psi_iy_i + log\, o_0\sum_{i=1}^ny_i + log\, o_{10}\sum_{i=1}^n \psi_i
\end{split}
\end{align}

where $\theta = \left(\psi, \pi_0, \pi_1, k\right)$, $o_0 = \pi_0 / (1-\pi_0)$, $o_1 = \pi_1 / (1-\pi_1)$ and $o_{10} = (1-\pi_1) / (1-\pi_0)$. That is, $o_0$ is the odds for unmarked samples, $o_1$ is the odds for marked samples, and $o_{10}$ is the quotient of negative probabilites for marked and unmarked samples. If we define $n_{pm}$ the number of positive and marked individuals, $n_p$ the number of positive and $n_m$ the number of marked individuals, we can finally write:

\begin{equation}\label{eq:loglike}
logP(y\mid X, \theta) = n\cdot log\left(1-\pi_0\right) + n_{pa}log\,\frac{o_1}{o_0} + n_alog\, o_0 + n_plog\, o_{10}
\end{equation}

The maximum likelihood estimate of probabilities $\pi_0$ and $\pi_1$, given $\psi$, are given by

\begin{equation}
\hat{\pi}_0 = \frac{n_a - n_{pa}}{n-n_p}, \qquad \hat{\pi}_1 = \frac{n_{pa}}{n_p}
\end{equation}

as expected.

\subsection{Priors}
Describing the boolean function $\psi$ using $\omega$, the set of the $m$ implicants involved in its DNF, we have that

\begin{equation}
\omega = \{S_1, S_2, ..., , S_m\}, \, S_i \in 2^{\{1,...,p\}}
\end{equation}

The dimension of the parametric space in this case is itself a parameter (namely $m$); inference in this kind of space is a known problem well studied in the literature of Bayesian model selection (see for example \cite{Stephens2000, Carlin1995, Green1995} or \cite{Sisson2005} for a review). The usual approach is to specify a prior for $\omega$ through the factorization

\begin{equation}
P(\omega) = \sum_m P\left(\{S_1, S_2, ..., S_m\} \mid m\right)P(m) = P(\{S_1, S_2, ..., S_{m^*}\}\mid m^*)P\left(m^*\right)
\end{equation}

since $P(\{S_1, S_2, ..., S_m\} \mid m) = 0$ if $m^* = \lvert \{S_i\} \rvert$, the cardinality of the set $\{S_i\}_i$, is different from $m$. The term $P\left(\{S_i\} \mid m\right)$ is a prior over all sets of $m$ conjunction terms (i.e., over the set of subsets of $\{1,...,p\}$ with cardinality $m$), and $P(m)$ is a prior on the number of terms. This prior can be used to control the sparsity of the function $\psi$, penalizing functions with many terms. 

The function $P(\{S_i\}_i \mid m)$ can be used to encode whatever prior information is available over the form of $\psi$, specially if this information involves more than one implicant simultaneously (for example, if it is believed that no primitive variable is involved in more than one implicant).

In the simplest of cases, assuming prior independence between all pairs of terms given $m$, we can write

\begin{equation}
P(\{S_1, S_2, ..., S_m\} \mid m) = \prod_{i=1}^m P(S_i \mid m)
\end{equation}

Now the function $P(S_i \mid m)$ is a distribution over the power set of $\{1,..,p\}$; again it is possible to factorize this as

\begin{equation}
P(S_i \mid m) = P(S_i \mid m, k_i)P(k_i \mid m)
\end{equation}

where $k_i$ is the cardinality of the set $S_i$ (i.e., the number of variables involved in implicant $i$ of $\psi$). The function $P(k_i \mid m)$ is a prior over this cardinality, and can also be used to control for sparsity (in this case, the number of terms in each implicant). 

This analysis leads to a two-level hierarchical prior

\begin{equation}
P(\{S_1,S_2,...,S_m\}) = P(\{S_1,S_2,...,S_m\} \mid m) = \prod_{i=1}^mP(S_i \mid k_i)P(k_i \mid m)P(m)
\end{equation}

where it is assumed that $S_i$ and $m$ are independent given $k_i$.

\subsubsection{$\pi_0$ and $\pi_1$}

The prior model for $\pi_0$ and $\pi_1$ can also be used to encode useful information about the problem at hand. 

Let us analyze first two degenerate cases: $P(\pi_0 = \pi_1) = 1$ and $P(\pi_0 = 0, \pi_1=1) = 1$. 

In the first situation there is no actual effect of the marker, and the function $\psi$ loses all relevance. The posterior depends on $p = \pi_0 = \pi_1$ only, and the problem becomes one of estimating a constant probability in a Binomial model. 

When $P(\pi_0 = 0, \pi_1 = 1) = 1$ the marker is known a priori to have a \textit{decisive}\footnote{We choose the term decisive and not causal, in order to avoid discussion of possible confounding effects. Such discussion is beyond the scope of this work.} effect (in the sense that $P(y)= 0$ for unmarked individuals, and $P(y)=1$ for marked ones). 

In this case the function $\psi$ becomes a perfect boolean classificator. Part of its truth table is known (the sample $X,y$); if $n$ is the number of rows of $X$, and assuming no pair of rows are equal, it is possible to determine $\psi$ up to a set of $2^{2^p-n}$ elements (all possible values of $y$ on the remaining $2^p-n$ combinations of $x$). If of course $n = 2^p$, $\psi$ is uniquely and immediately determined. Otherwise, techniques for obtaining an optimal representation of $\psi$ exist (see for example the shadow clustering method of \cite{Muselli2005}), where optimality is usually a balance between the number of terms and the total number of literals appearing in the representation. 

In most practical situations, however, there will be less certainty about the effect size. In the extreme case of no information, a uniform prior in the rectangle $[0,1] \times [0,1]$ can be adopted.

%Finally, since the likelihood function in \ref{eq:like} factorizes as the product of two Bernoulli likelihoods, one for each probability parameter, the conjugate prior for $\pi_0$ and $\pi_1$ is a product of two Beta distributions $\mathcal{B}(a_0,b_0) \cdot \mathcal{B}(a_1,b_1)$. In this case the conditional distribution $P(\pi_0, \pi_1 \mid S, X, k)$ is again the product of two Beta distributions with parameters $a_0 + (n_p - n_{pm})$, $b_0 + (n - n_m) - (n_p - n_{pm})$, $a_1 + n_{pm}$, $b_1 + n_m - n_{pm}$. These priors can be used to propose Gibbs steps for $\pi_0$ and $\pi_1$ in the MCMC algorithm.

\section{Posterior inference}\label{sec:post}

As already noticed above, the space where (the DNF representation of) $\psi$ lies is mixed, i.e., it involves the cartesian product of discrete and continuous spaces. Moreover, the number of terms in the DNF is itself a parameter. 

Inference in these kinds of models can be done analytically, by adopting conjugate distributions (\cite{Consonni1995}), or numerically, which is usually done with Reversible Jump Markov Chain algorithms (\cite{Green1995}).

We start by analyzing the case $m=1$, i.e., $\psi$ can be represented by a single conjunction.

\subsection{Multivariate markers}

Suppose now for simplicity that $m=1$, i.e., the function $\psi$ is a single conjunction and $\omega = S \subset \{1,...,p\}$, i.e., in this case the $\psi$ function is completely defined by a subset of $\{1,...,p\}$. We call this conjunction a \emph{multivariate marker}. 

The prior for $S$ can then be factorized as

\begin{equation}
P(S) = \sum_k P(S\mid k)P(k) = P\left(S\mid k_S \right)P\left(k_S\right)
\end{equation}

Here $k_S =  \lvert S \rvert$.

The prior $P(k)$ must be chosen to reflect our knowledge over the size (number of primitive variables) of the interaction term. 

A value $k=0$ is equivalent to $\pi_1 = \pi_0$, modelling the case where there is no relation whatsoever between explanatory ($X$) variables and the outcome $y$. A value $k=1$ represents the case where a single variable carries all the effect. Values of $k$ greater than $1$ represent the actual multivariable marker case.

It is possible to choose priors that assign little or no mass to this two points, imposing \textit{a priori} that there must be a marker and that it must be a multivariable one. It can be useful in situations where marginal effects of single variable markers are not expected. 

Of course, whatever is the case it must hold that $P(k \le p) = 1$. Given only this set of conditions ($0 \le k \le p$ with probability one, $k$ integer values), the most conservative (i.e. maximum entropy) prior would be the uniform on $\{0,...,p\}$. In most practical cases, however, one expects that $k$ will be much smaller than $p$; the prior can then be adjusted accordingly.

\subsubsection{Posterior}

The posterior por $S$ can now be obtained through Bayes' theorem

\begin{equation}
P(S \mid y, X, \pi_0,\pi_1,k) = \frac{P(y \mid S, X, \pi_0, \pi_1, k)P(S\mid X, \pi_0, \pi_1, k)}{P(y\mid X, \pi_0, \pi_1, k)}
\end{equation}

We assume that $P(S\mid X,\pi_0,\pi_1,k) =P(S\mid k)P(k)P(\pi_0,\pi_1)$, that is, we assume prior independence between the marker (both its size and contents) and the effect size.

There is however one particular situation that deserves attention, which is the case when there is information available about the relative frequency of positives in the entire population, i.e., information about $P(y)$. According to the above model, the following identity holds

\begin{align}
\rho = P(y)& = P(y \mid \psi(x) = 1)P(\psi(x) = 1) + P(y \mid \psi(x) = 0)P(\psi(x) = 0) \\
& = \pi_1P(\psi(x) = 1) + \pi_0P(\psi(x) = 0) \\
& = \pi_1\theta + \pi_0(1-\theta)
\end{align}

That is, the marginal probability $P(y)$ is a convex combination of $\pi_0$ and $\pi_1$, with coefficients $\theta = P(\psi(X)=1)$ and $1 - \theta$, and in this case prior independence between the marker and the probabilities is lost (conditionally on $\rho$).

The posterior then takes the form

\begin{equation}\label{eq:post}
P(S\mid y,X,\pi_0,\pi_1,k) \propto P(y \mid S,X,\pi_0,\pi_1,k)P(S\mid k)P(k)P(\pi_0, \pi_1)
\end{equation}

where $S$ is a subset of $\{1,...,p\}$ with cardinality $k$. 

\subsubsection{Posterior inference}\label{sec:mcmc}

In the simple case $m=1$ it is possible to avoid the complications of defining Markov chains that transverse spaces of different dimensions. If the parameter $S$ is defined as a collection of indices $1 \leq i \leq p$, then it would live in a space of dimension $k$, and there would be the problem of different dimensions. Another possible parameterization, however, is in terms of a binary vector $\xi \in \{0,1\}^p$, where each entry of $\xi$ is $1$ if the corresponding index is in $S$, and $0$ otherwise. This is a parameterization typical of Bayesian model selection methods such as spike-and-slab regression (\cite{Ishwaran2005}).

With this parameterization, to obtain samples from model \ref{eq:post} we define a Markov chain with state $x_t = (\xi, p_0, p_1) \in \{0,1\}^p \times [0,1]\times [0,1]$. 

The transition kernel of the chain allows two kinds of moves:

\begin{enumerate}
\item \textbf{Switch} one component of $\xi$;
\item \textbf{Sample} $(\pi_0, \pi_1)$.
\end{enumerate}

Adopting two independent Beta priors for $(\pi_0,\pi_1)$, the conditional posterior on $(\pi_0,\pi_1)$ will also be the product of two Betas, and the second step is made into a Gibbs step.

For the first step, a randow walk, Hastings procedure is adopted: a candidate move is obtained by sampling $j \sim Unif(\{1,...,p\})$, and switching the corresponding value $\xi_j \leftarrow 1 - \xi_j$. If this step is accepted, the value of $k$ is adjusted accordingly.

\subsection{Multiple multivariate markers}

When $m > 1$, i.e., when the DNF of function $\psi$ depends on more than one conjunction term, the problem of posterior inference becomes more involving.

A first approach might be to extend the idea of reparameterizing the space in terms of boolean vectors: since the power set of $\mathcal{U}=\{1,...,p\}$ has cardinality $2^p$, it is possible to define a transformation from $2^{\mathcal{U}}$ to the integers smaller than $2^p$. By doing so, the parametric space becomes the space of boolean vectors with dimension $p$ or, equivalently, the space of integers from $0$ to $2^p-1$. The on-bits of $\psi \in \{0,1\}^{2^p}$ would point to the integer representation of each multivariate marker.

For instance, suppose for simplicity that $p = 3$. There are $8$ possible conjunction terms involving the three variables. Assuming the usual binary representation of an integer, it is possible to map the empty set to $0$, the set including only the third variable to $1$, the set including the second variable only to $2$ and so on. Now the function $\psi$ can be represented by a binary string of length $8$; suppose $\psi = 00011000$. This can be transformed to the set representation of $\psi$ by first forming the set of indices of the conjunction terms, which in this case is given by $\{3,4\}$, and finally by transforming each integer to its $p$-digit binary representation, obtaining $\psi = \{\{011\}, \{100\}\}$, or $\psi(x) = x_1x_2 + x_0$.

The main issues with this idea are the prohibitive memory requirements if $p$ becomes too large, and the computational cost of applying the function $\phi$, encoded as an integer, to the variables $X$. It is in principle possible, however, to imagine a clever way to implement such an algorithm by resorting to low-level languages working directly on binary values as stored in memory. This picture is illustrative nevertheless, as it gives a good feeling about the size of the space in which $\psi$ lies. 

The set representation of $\psi$ writes it as a point $\psi \in 2^{\mathcal{U}} \times 2^{\mathcal{U}} \times ... \times 2^{\mathcal{U}} = \left[2^{\mathcal{U}}\right]^m$ where each coordinate represents a conjunction term. In this parameterization, posterior inference will necessarily involve transdimensional methods, i.e., methods that are capable of searching in spaces formed as the union of subspaces with varying dimension. 

The Reversible Jump MCMC of Green (\cite{Green1995, Hastie2012}) is one such method, which has already proven to be effective in many problems. The idea is to build an ergodic Markov Chain on the union space, assuring detailed balance for each move type. The challenging situation is, as expected, to assure the balance between moves between spaces with different dimensions.

Green achieves this balance by proposing to augment both spaces in such a way that the augmented spaces have equal dimensions, and then defining a diffeomorphism between these spaces (for details, please see \cite{Hastie2012} and references therein). 

\subsection{RJMCMC for boolean function learning}

A boolean function whose DNF consists of a single conjunction term does not pose any particular technical difficulties for posterior estimation, as seen in the last sections. When the number of conjunction terms is allowed to vary, however, it is necessary to consider balance between transdimensional moves in the design of the MCMC algorithm.

Conisder a Markov chain with only two types of transdimensional moves: a \textit{birth}, that increases the space dimension from $m-1$ to $m$, and a \textit{death} that decreases the space dimension from $m$ to $m-1$.

Suppose then that $\psi \in \left[2^{\mathcal{U}}\right]^m$. There are many different ways to transform this $\psi$ to a lower dimensional version; probably the simplest would be to randomly delete one coordinate of $\psi$, with the reverse move being given by a random selection from $2^\mathcal{U}$. This move, however, would lead to chains with slow mixing time, for the deletion move would throw away information (since the deleted coordinate is probably located in a region of higher posterior mass).

One could adopt $\textit{split and merge}$ move types instead: selecting two coordinates of $\psi$ and merging them into one preserve some of the information already accumulated by the chain.

The question is then how to perform the split, and, more importantly, how to define the reverse merging move. Green's theory (and most applications of RJMCMC in the literature) is originally built to work on the product of continuous spaces such as $\mathbb{R}^{n_1} \times \mathbb{R}^{n_2} \times ...$. In this setting, when moving from $\mathbb{R}^n$ to $\mathbb{R}^m$ with $n < m$, say, he proposes augmenting the first space with $m-n$ continuous random variables, and defining a diffeomorphism between $\mathbb{R}^n \times \mathbb{R}^{m-n}$ and $\mathbb{R}^m$; the differentiability requirement is needed in order to obtain acceptance probabilites which depend on the Jacobian of the transformation.

The current situation is somewhat different, since the spaces with varying dimensions are discrete. It is possible, however, to adapt the methodology of Green to this problem, as shown below.

\subsubsection{RJMCMC for set spaces}

Define a Markov chain with state given by $s_t = \left(p_0, p_1, f_m\right)$, where the subscript $m$ indicates that the PDNF of the boolean function $f$ is written using $m$ conjunction terms. That is, $f \in \left[2^\mathcal{U}\right]^m$. The transdimensional steps we propose are: adding a new conjunction term ($f_m \to f_{m+1}$), or removing a conjunction term ($f_m \to f_{m-1}$). 

What is needed next is then \textit{a)} a way to augment the lower dimensional space, which then becomes $\left[2^\mathcal{U}\right]^{m-1} \times 2^\mathcal{U}$ ; \textit{b)} a transformation between the original $m$ dimensional space, and the augmented $(m-1) + 1$ dimensional space. 

In the original formulation of RJMCMC, this transformation must have an inverse, and also be a diffeomorphism, in order that the Jacobian can be calculated. This conditions are necessary in the calculation of the acceptance probability for transdimensional moves. 

Denoting the full posterior by $\pi(\theta)$ and considering only the moves between $m$ and $m-1$ spaces (either way), the intra-move detailed balance condition can be written as

\begin{equation}\label{eq:detailed}
\sum_{f \in A, g \in B}\pi(f)P(f, g) = \sum_{f \in A, g \in B}\pi(g)P(g, f)
\end{equation}

for all sets $A \subset \left[2^{\mathcal{U}}\right]^m, B \subset \left[2^{\mathcal{U}}\right]^{m-1}$, and where $P(f,g)$ represents the probability of a transition between the states $f$ and $g$ (here we ommited dependence on $p_0$ and $p_1$ for simplicity of notation). This transition probability is the product of the candidate distribution (which might depend on the current state) and an acceptance probability.

Now define the transformation $h:\left[2^\mathcal{U}\right]^2 \to 2^\mathcal{U}$ as $h(u,v) = u \triangle v$, the symmetric difference of sets $u$ and $v$. The choice of this transformation is motivated by the fact that the symmetric difference is the only set operation that can be inverted, in the sense that $(u\triangle v)\triangle v = u$ and $(u\triangle v)\triangle u = v$. 

Using this transformation, it is possible to define forward and backward moves as following:

\begin{enumerate}
\item $m\to m-1$: select two elements of $f$, $u_1, u_2$, and replace them by $u = u_1 \triangle u_2$;
\item $m-1 \to m$: select one element $u$ of $g$, one element $w \in 2^\mathcal{U}$, and replace $u$ by $(w, u\triangle w)$.
\end{enumerate}

Consider the case $m=2$. Since the transformation $h$ defines a bijection from $\left[2^\mathcal{U}\right]^2$ into itself, the detailed balance requires equality between the transition probabilities 

\begin{equation}
P(f,g) = \alpha(f,g)P(f) = \alpha(g,f)2^{-p}P(g) = P(g,f)
\end{equation}

Then for each $f =  (u_1, u_2)$, and each $g = u$ such that $u = u_1 \triangle u_2$, the balance condition becomes

\begin{equation}
\pi(f)\alpha (f,g) = \pi(g)\alpha(g,f)P(u_1)
\end{equation}

where $P(u_1)$ is the candidate mass at point $u_1$, and in this case the acceptance probability

\begin{equation}\label{eq:acc}
\alpha(f,g) = min\left\{1, \frac{\pi(g)P(u_1)}{\pi(f)2^p}\right\}
\end{equation}

can be used.

The Markov chain thus defined will be ergodic, and sampling from it will eventually provide samples from the posterior distribution of our model.

The issue, however, is that the symmetric difference might not be a good way to traverse the parameter space, because at each death step we loose information about the intersection of the two selected terms. Also, the acceptance probability and mixing properties of this chain will be far from ideal, since the birth move will select any conjunction term from the entire set $2^p$, and this will be likely to take the chain to states in regions with lower posterior mass.

These observations, combined with the fact that the state space of this chain is very large, motivates us to propose a different algorithm for posterior optimization in the case of multiple multivariate markers. Instead of defining a Markov chain and aiming at obtaining samples from the full posterior, we propose a stochastic search algorithm to optimize the posterior.

\subsection{Stochastic search}

Stochastic search algorithms are optimization procedures that take random steps through the set of feasible points. They're widely adopted in combinatorial problems, where there is no gradient available.

For the optimization of the posterior for the multiple markers case we propose a simulated annealing algorithm, inspired by the MCMC structure from last section.

Simulated annealing is a well known algorithm that links the Metropolis method with combinatorial optimization\cite{Kirkpatrick1983}. Given a representation of a point in  the feasible space, a probabilistic dynamic is the defined to allow the algorithm to move between feasible points and explore the configuration space. As in classical Metropolis algorithm, a move to a point with higher posterior (compared to the current point) is always accepted, and a move to a point with lower posterior is randomly accepted with a given probability. In the simulated annealing algorithm, this probability decays over time (i.e. over iterations). In the early stages of the algorithm, then, the configuration space is quickly traversed and explored, and as the acceptance probability falls (\textit{cooling}) the algorithm will converge to a locally optimal point. 

\subsubsection{Simulated annealing for boolean function learning}

The most important component of a simulated annealing algorithm is the dynamics, i.e., the moves that allow the algorithm to explore the space. For our model of the boolean function estimation problem, there are three groups of moves that must be included: \textit{a)} changing $p_0$ and $p_1$; \textit{b)} changing one multivariate markers; \textit{c)} adding a new marker or deleting a current marker.

The key for an efficient algorithm lies in the design of such moves. 

For the first group, the (conditionally) conjugate Beta distribution provides an obvious strategy to generate candidate moves. For changing a current marker, the move must consist in choosing one one marker from the current set, and then including (deleting) a variable in (from) the marker. The algorithm we propose will define two moves: one, selecting a marker with probability proportional to its current size, and then randomly removing a variable from the marker. Two, selecting a marker with probability inversely proportional to its current size, and then including a random new variable in the marker.

The third group is the most challenging, as it involves the change of dimension of the current point. For the death move (removing a term from the disjunction) there are one obvious way to proceed, which is to randomly select one marker and thoroughly remove it from the set; another possible approach would be to randomly select a pair of markers, and replace them by ther intersection. This second strategy helps the algorithm to stay in high posterior mass regions, by preserving some of the information condensated in the current point.

Finally, the birth move is the most challenging, since there is no obvious weay to select a good candidate for the new marker; selecting randomly from the entire set of possible markers will be very unefficient, for it will most likely select a marker that is (very) far from the high posterior mass regions. 

One possible approach is to select a new marker by taking one row of $X$ as the new marker. This strategy guarantees that the algorithm does not wander too much in the configuration space, stopping it for example to include a marker involving a conjunction that is not satisfied by any individual in the dataset.

Our final algorithm, then, consists of $7$ move types:

\begin{enumerate}
\item Sample $p_0$ from the conditional posterior (Beta);
\item Sample $p_1$ from the conditional posterior (Beta);
\item Select one marker with probability proportional to marker size, flip one bit off;
\item Select one marker with probability inversely proportional to marker size, flip one bit on;
\item Include a new marker, taken from the set of affected individuals in the sample;
\item Select a pair of markers, replace them by their intersection;
\item Select one marker and remove it from the current function.

\end{enumerate}

Moves are randomly selected at the beginning of each iteration. In this version of the algorithm the selection probabilities remain fixed, but it is possible to develop an adaptive scheme for selecting the move type. 

If the posterior is increased, the move is accepted with certainty; otherwise, it will be randomly accepted wtih probability

\begin{equation*}
min\left\lbrace 1, \frac{P(s_{cand} \mid X, y, \theta) }{P(s_{curr} \mid X, y, \theta)}\cdot \lambda_i\right\rbrace
\end{equation*}

where $P(\cdot \mid X, y, \theta)$ is the posterior and $\lambda_i$ is a decreasing sequence of positive numbers. During the initial steps, $\lambda_i$ is used to increase the acceptance probability, even when the proposed state has a much lower posterior value. This is meant to facilitate the exploration of the parameter space in the early stages of the algorithm. As $\lambda_i$ decreases, moves that take the process to states with lower posterior probability will be more and more unlikely to be accepted. A cooling schedule for $\lambda_i$ must be adopted; one common choice is to take $\lambda_{i+1} = \lambda_i \cdot \rho$ with $0 < \rho < 1$ the cooling factor.

Since the most difficult moves in this setup are the inclusion of new terms in the current marker, we apply the modification of the acceptance probability for this move only. 

\section{Experimental results}\label{sec:results}

\subsection{Simulated data}

To conduct a first test of the proposed algorithms, we simulate binary data $X,y$ in the following way:

\begin{enumerate}
\item[] Given number of samples $n$, number of variables $p$, number $m$ of conjunction terms in $\psi$, size of each marker $k_1,...,k_m$, proportion of marked individuals in the sample $p_{mark}$, and probabilities $\pi_0, \pi_1$:
\item[1] Define $\psi = x_1x_2...x_{k_1} + x_{k_1+1}...x_{k_1 + k_2} + ...$;
\item[2] Generate $\beta \in [0,1]^p$ from a uniform and independent distribution;
\item[3] \textbf{For} $i=1,...,n$:
\begin{enumerate}
\item[3.1] Generate $x_i$ from independent Beta distributions with probabilities given by $\beta$;
\item[3.2] Calculate $\psi(x_i) = j$;
\item[3.3] \textbf{If} $j=1$, generate $y_i \sim \mathcal{B}(\pi_1)$; otherwise generate $y_i \sim \mathcal{B}(\pi_0)$.
\end{enumerate} 
\end{enumerate}

This procedure emulates random sampling from a population. The final proportion of affected individuals in this sample will depend on the size of the marker and on $\pi_0$ and $\pi_1$.

\begin{figure}[H]
\centering
\includegraphics[width=15cm]{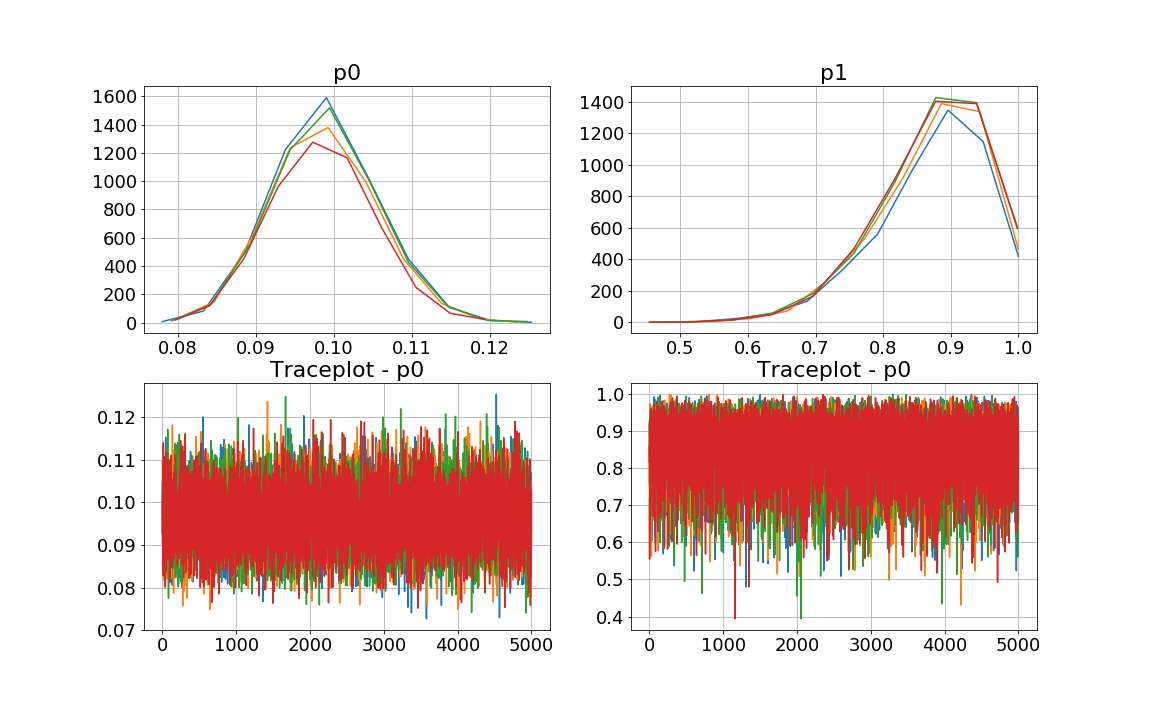}
\caption{Posterior histograms and traceplots for $\pi_0$ and $\pi_1$}
\label{fig:fig1}
\end{figure}

\begin{figure}[H]
\centering0
\includegraphics[width=15cm]{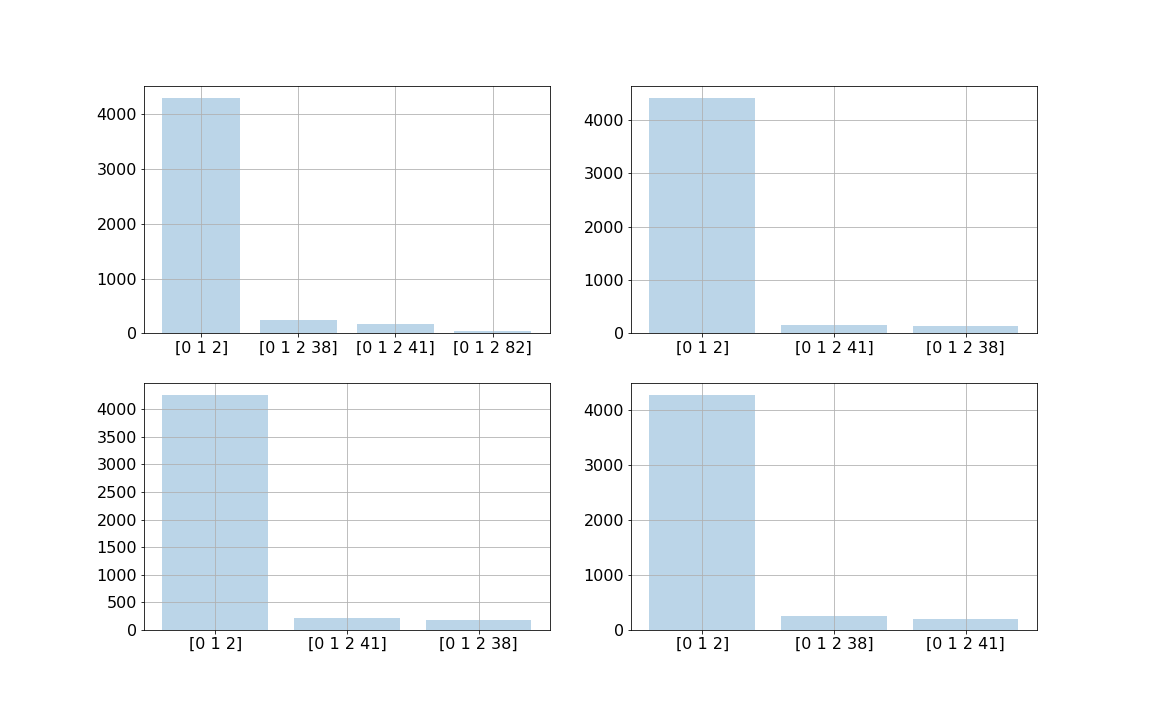}
\caption{Posterior histograms for $\psi$}
\label{fig:fig2}
\end{figure}

As a first test, we use a $\psi$ function with a single conjunction term, and take $N = 2000, p = 100, k = 3, \pi_0 = 0.1, \pi_1 0.9$. We ran four parallel chains starting with markers selected randomly from the rows of $X$. After running each chain for $10,000$ iterations, we take the final $5,000$ generated points as the sample from the posterior. The posterior histograms and traceplots for $\pi_0$ and $\pi_1$ are shown in figure \ref{fig:fig1}, and the posterior histograms for $\psi$ are shown in figure \ref{fig:fig2}. All runs in this section use a non-informative prior for $\psi$, and $\beta(1,1)$ independent priors for both $\pi_0$ and $\pi_1$.

The chains converge and correctly attributes high posterior mass to the actual conjunction term in function $\psi$. 

Of course in this toy example both the effect and sample sizes are large. For fainter effects and keeping the sample size constant, the convergence will be slower and the posterior less concentrated. 

The sample size has an important effect in this setup, specially because of the random sampling schema, where depending on the size of the marker the marked individuals will appear very rarily. In these cases, it might be best to adopt an informative sampling scheme (for instance sampling affected individuals with higher probability), and adjust the model accordingly.

\subsubsection{Multiple markers}

Next we simulate a sample in the same conditions as the first one, but now with $\psi$ consisting of two terms, boith with size $2$. Running the MCMC for the single marker case on this data we find that the chains converge to the single marker which happens to be more frequent in the database. 

To apply the simulated annealing algorithm we take $ln\left(\lambda_0\right) = 1000$ and $\rho = 0.9$. This forces the algorithm to accept a lot of increase dimension moves. Our tests indicate that high values of these parameter are necessary for the algorithm to be able to effectively explore higher dimensions. 

To choose an initial point for each algorithm we first choose the number $m_0$ of markers, an then select $m$ individuals from the sample, taking their corresponding rows in $X$ to be the conjunction terms.

An algorithm with good properties would be insensitive to the inital point. However, given the combinatorial nature of the problem, we suggest to set $m_0$ at high values. In this way the death steps are more effective and the algorithm will have a superior performance.

The hyperparameters involved in the prior for the function $\psi$ are also important. If running with an entirely non-informative prior for $\psi$, the algorithm will likely increase the size of $\psi$ (both its number of terms and the size of each term), trying to capture as many affected individuals as it can. In the extreme case, if there is a perfect monotone boolean discriminator $\psi$ for this dataset, the algorithm with non-informative priors will tend to move towards it (since it is the global maximum point of the likelihood).

To verify the effect of the hyperparameters, we run the algorithm on the simulated data using $m_0 = 1, m_0 = 10$ and $m_0 = 20$; for each value of $m_0$ we run the algorithm from $L$ parallel starting points, such that $m_0 \cdot L = 20$. We repeat these runs three times: one with a completely non-informative prior, one using independent Poisson priors with $\theta = 10$ for the size of each conjunction term and a non-informative prior for $m$ (the number of markers), and finally using the Poisson priors for the size and a geometric prior with $p = 0.5$ for $m$. Results appear in table \ref{table:tbl1}. A total number of $100$ steps is performed in each trial.

\begin{table}[H]\label{table:tbl1}
\centering
\begin{tabular}{cccccccccc}
\toprule
 nstart &  nchains &  $\theta$ &  $p_{geom}$ &  $\pi_0$ &   $\pi_1$ &    $m$ &  $\sum k$ &  term 1 &  term 2 \\
\midrule
      1 &       20 &    0.0 &    0.0 & 0.34 & 0.73 &   2.4 &   45.9 &    0.2 &    0.0 \\
      1 &       20 &   10.0 &    0.0 & 0.31 & 0.67 &   1.4 &   16.6 &    0.2 &    0.0 \\
      1 &       20 &   10.0 &    0.5 & 0.38 & 0.68 &   1.3 &   10.4 &    0.1 &    0.0 \\
     10 &        2 &    0.0 &    0.0 & 0.38 & 0.74 &  13.0 &  516.5 &    0.0 &    0.0 \\
     10 &        2 &   10.0 &    0.0 & 0.39 & 0.64 &   8.4 &  272.7 &    0.0 &    0.0 \\
     10 &        2 &   10.0 &    0.5 & 0.38 & 0.58 &   7.2 &  212.7 &    0.0 &    0.0 \\
     20 &        1 &    0.0 &    0.0 & 0.39 & 0.59 &  20.1 &  826.0 &    0.0 &    0.0 \\
     20 &        1 &   10.0 &    0.0 & 0.39 & 0.53 &  19.9 &  803.6 &    0.0 &    0.0 \\
     20 &        1 &   10.0 &    0.5 & 0.39 & 0.49 &  20.1 &  806.6 &    0.0 &    0.0 \\
\bottomrule
\end{tabular}
\caption{Simulated annealing on synthetic data: $10$ independent datasets, mean of optimal values of $\pi_0$ and $\pi_1$, size of final marker ($m$), total number of atomic terms ($\sum k$), and proportion of runs which detected term 1 and term 2, with $N = 1000$}
\end{table}

The small number of steps prevented the algorithm from finding the correct function $\psi$ in all cases; however, the effect of the prior hyperparameters is evident. Also it becomes aparent that it is best to start many parallel chains with a single term in the starting point for the function $\psi$, than starting a single chain with a starting $\psi$ with many terms. To analyze this further, we rerun the simulations, now adopting $\theta = 10$ and $p_{geom} = 0.5$ throughout, but increasing the number of steps to $10,000$ in each trial. Results appear in table \ref{table:tbl2}. 

\begin{table}[H]\label{table:tbl2}
\centering
\begin{tabular}{cccccccc}
\toprule
 nstart &  nchains &   $\pi_0$ &   $\pi_1$ &   $m$ &  $\sum k$ &  term 1 &  term 2 \\
\midrule
      1 &        1 & 0.24 & 0.90 &  1.1 &    2.2 &    0.6 &    0.5 \\
      1 &        2 & 0.16 & 0.90 &  1.6 &    3.2 &    1.0 &    0.6 \\
      1 &       20 & 0.10 & 0.90 &  2.1 &    4.7 &    1.0 &    1.0 \\
     10 &        1 & 0.16 & 0.90 &  1.6 &    3.2 &    0.9 &    0.7 \\
     10 &        2 & 0.16 & 0.90 &  1.6 &    3.2 &    0.7 &    0.9 \\
     10 &       20 & 0.12 & 0.90 &  1.9 &    3.8 &    0.9 &    1.0 \\
     20 &        1 & 0.20 & 0.90 &  1.4 &    2.8 &    0.6 &    0.8 \\
     20 &        2 & 0.14 & 0.90 &  1.7 &    3.4 &    0.9 &    0.8 \\
     20 &       20 & 0.10 & 0.90 &  2.1 &    4.5 &    1.0 &    1.0 \\
\bottomrule
\end{tabular}
\caption{Simulated annealing on synthetic data ($10$ independent simulated datasets); average estimated values for $\pi_0$ and $\pi_1$, size of final marker ($m$), total number of atomic terms ($\sum k$), and proportion of runs that correctly included each marker. $N = 1000$}
\end{table}

The size of the initial point for function $\psi$ is not as important as the number of parallel chains. This is to be expected since the algorithm is built to favour transdimensional steps at the early stages. With a sufficient number of parallel points, $29$ out of $30$ runs converged to the correct point.

As a last test, we rerun the simulated annealing algorithm with $m_0=1$ and $20$ parallel chains, now varying the sample size of the simulated datasets; we simulate $100, 250, 500$ points on the same conditions. The results are in table \ref{table:tbl3}.

\begin{table}[H]\label{table:tbl3}
\centering
\begin{tabular}{ccccccc}
\toprule
   $N$ &  $\pi_0$ &  $\pi_1$ &    $m$ &  $\sum k$ &  term 1 &  term 2 \\
\midrule
 100 & 0.23 & 0.88 &  1.1 &    2.2 &    0.6 &    0.5 \\
 250 & 0.18 & 0.90 &  1.5 &    3.0 &    0.7 &    0.8 \\
 500 & 0.11 & 0.88 &  1.9 &    3.8 &    0.9 &    1.0 \\
\bottomrule
\end{tabular}
\caption{Simulated annealing on synthetic data ($10$ independent simulated datasets); average estimated values for $\pi_0$ and $\pi_1$, size of final marker ($m$), total number of atomic terms ($\sum k$), and proportion of runs that correctly included each marker, for $N \in \{100, 250, 500\}$}
\end{table}

For a sample size of $N=500$ the algorithm correctly identified the function $\psi$ in $9$ out of $10$ simulated datasets. For smaller sample sizes, the performance is worse, but even with $N=100$ one of the correct terms in $\psi$ was correctly identified half of the time.

\subsection{Mushroom dataset}\label{sec:app}

To test the simulated annealing algorithm in a real dataset, we pick the Mushroom dataset from UCI repository\footnote{Available at \url{https://archive.ics.uci.edu/ml/datasets/mushroom}}. This is a dataset for a binary classification problem with categorical explanatory variables, and our proposed method is well suited to the dataset.

There are $8124$ individuals with $21$ categorical attributes (\textit{veil-type} is removed as all individuals have the same value for this variable). After converting the multicategory variables to binary (dummy) variables, the design matrix has $116$ columns.

To select the hyperparameters of our model, we use a cross-validation strategy, splitting the data in training and testing subsets with $4062$ individuals each (i.e., we use $50\%$ of the data to learn the boolean function, and test the function learned on the other $50\%$).

Applying a grid search procedure for the hyperparameters (Poisson priors for the number of atomic variables in each conjunction term with $\theta \in \{2, 5, 10, 30\}$ and a geometric prior for the number of conjunction terms with $p_{geom} \in \{0.1, 0.5, 0.9\}$, we find that most combinations of hyperparameters converge to boolean functions with very low generalization error as estimated by the cross-validation procedure. Table \ref{table:tbl4} shows the average AUC of the boolean function classifier for a $10$ fold repetitions of the sample split. The table also shows the average size of the estimated function $m$ and the average number of atomic variables in the estimated function, $\sum k$. Each run of the algorithm takes on average $2.6$ minutes on a desltop with $8$ CPUs and $24$ $Gb$ of RAM, running Fedora 27.

\begin{table}[H]\label{table:tbl4}
\centering
\begin{tabular}{ccccc}
\toprule
 $\theta$ &  $p_{geo}$ &  $m$ &  $\sum k$ & AUC \\
\midrule
     2 &   0.1 &  6.9 &   13.4 & 0.9993 \\
     2 &   0.5 &  6.9 &   13.3 & 0.9992 \\
     2 &   0.9 &  7.0 &   13.3 & 0.9993 \\
     5 &   0.1 &  6.7 &   14.9 & 0.9999 \\
     5 &   0.5 &  6.5 &   14.4 & 0.9991 \\
     5 &   0.9 &  6.7 &   15.7 & 0.9995 \\
    10 &   0.1 &  6.6 &   19.9 & 0.9994 \\
    10 &   0.5 &  6.3 &   17.2 & 0.9988 \\
    10 &   0.9 &  6.1 &   16.4 & 0.9987 \\
    30 &   0.1 &  6.5 &   32.1 & 0.9998 \\
    30 &   0.5 &  6.5 &   29.4 & 0.9999 \\
    30 &   0.9 &  5.8 &   24.6 & 0.9985 \\
\bottomrule
\end{tabular}
\caption{Cross-validation results for the Mushroom dataset using half of the sample as training set and hal as test set, repeated $10$ times.}
\end{table}

On this dataset the effect of the geometric prior is weaker than the effect of the Poisson prior for the total number of atomic terms. The best AUC where achieved by $\theta = 30, p_{geo} = 0.5$ and $\theta = 5, p_{geo} = 0.1$, with AUC of $0.999898$ for both.

Table \ref{table:tbl5} shows the results for each of the $10$ repetitions when $\theta = 5$ and $p_{geo} = 0.1$. 

\begin{table}[H]\label{table:tbl5}
\centering
\begin{tabular}{cccc}
\toprule
Run &  $m$ &  $\sum k$ & AUC \\
\midrule
0 &  6 &      14 & 0.9990 \\
1 &  7 &      22 & 1.0000 \\
2 &  7 &      14 & 1.0000 \\
3 &  6 &      15 & 1.0000 \\
4 &  7 &      14 & 1.0000 \\
5 &  6 &      14 & 1.0000 \\
6 &  7 &      15 & 1.0000 \\
7 &  7 &      13 & 1.0000 \\
8 &  7 &      15 & 1.0000 \\
9 &  7 &      13 & 1.0000 \\
\bottomrule
\end{tabular}
\caption{Cross-validation results for the Mushroom dataset using half of the sample as training set and hal as test set, repeated $10$ times with fixed hyperparameters}
\end{table}

In $9$ out of $10$ runs the algorithm identified a boolean function with $AUC = 1$. The functions obtained are not the same, which indicates that there are many logical rules that can be used to describe a poisonous mushroom given the features in the dataset. 

The logical expressions derived from the boolean function estimated in each case appear in appendix \ref{sec:appA}. 

\section{Conclusion}\label{sec:conclusion}

The problem of binary classification can be seen as a problem of estimating a partition on Bernoulli data. When the explanatory variables are all categorical, the problem can be cast in the language of boolean function estimation.

In this work we propose a Bayesian algorithm to estimate Bernoulli partitions based on boolean functions. Despite the complexity of the problem, the probabilistic methods and in particular the proposed simulated annealing algorithm show promising results in identifying patterns that have good classification performance and low generalization error. Also this method provides classificators that are immediately interpretable as logical rules, which can be useful in many applications.

The implementation of the algorithms studied in this paper, along with iPython notebooks that can be used to replicate our results are available on \url{http://github.com/paulohubert/babool}.

\section*{Acknowledgments}

This work is made possible by a generous funding from \textit{Instituto Paulo Gontijo}, a non-profit brazilian organization dedicated to the study of Lateral Amyotrophic Sclerosis. The author is deeply grateful for their support.

\newpage
\appendix
\section{Logical expressions estimated for the Mushroom dataset}\label{sec:appA}

\begin{enumerate}
\item (stalk-root = c AND stalk-surface-below-ring = y) OR (cap-shape = x AND odor = c AND ring-type = p) OR (odor = p AND stalk-root = e) OR (odor = f AND veil-color = w) OR (stalk-surface-below-ring = s AND spore-print-color = r) OR (gill-size = n AND stalk-root = ? AND ring-type = e)
\item (bruises = f AND stalk-shape = e AND ring-type = n) OR (odor = f AND veil-color = w AND ring-number = o) OR (bruises = t AND gill-size = n AND stalk-shape = e) OR (gill-spacing = c AND stalk-color-above-ring = w AND spore-print-color = r) OR (cap-color = y AND gill-spacing = w AND stalk-color-below-ring = y) OR (odor = c AND gill-size = n AND stalk-surface-above-ring = s) OR (gill-spacing = c AND ring-type = e AND spore-print-color = w AND population = v)
\item (stalk-surface-above-ring = k AND habitat = d) OR (bruises = t AND gill-size = n AND stalk-shape = e) OR (odor = f AND veil-color = w) OR (stalk-surface-above-ring = s AND spore-print-color = r) OR (stalk-surface-below-ring = y AND ring-type = e AND habitat = l) OR (gill-color = b) OR (odor = c)
\item (stalk-root = ? AND ring-number = o AND ring-type = e) OR (stalk-surface-below-ring = s AND spore-print-color = r) OR (odor = f AND gill-attachment = f AND veil-color = w) OR (odor = c) OR (bruises = t AND gill-size = n AND stalk-shape = e) OR (bruises = f AND stalk-root = c AND spore-print-color = w)
\item (stalk-surface-below-ring = s AND spore-print-color = r) OR (gill-spacing = w AND population = c AND habitat = l) OR (odor = c) OR (odor = f AND veil-color = w) OR (gill-spacing = c AND stalk-surface-above-ring = k) OR (gill-color = b) OR (odor = p AND stalk-shape = e AND stalk-surface-below-ring = s)
\item (bruises = f AND gill-spacing = c AND ring-type = e) OR (stalk-root = c AND spore-print-color = w) OR (bruises = t AND gill-size = n AND stalk-shape = e) OR (odor = f AND veil-color = w) OR (stalk-surface-above-ring = s AND spore-print-color = r) OR (odor = c AND stalk-root = b)
\item (gill-size = n AND stalk-root = ? AND spore-print-color = w) OR (stalk-shape = e AND stalk-surface-below-ring = s AND habitat = d) OR (gill-spacing = w AND population = c) OR (odor = f) OR (spore-print-color = r) OR (odor = p AND gill-attachment = f AND stalk-color-above-ring = w) OR (odor = m AND stalk-color-below-ring = c)
\item (gill-spacing = c AND gill-size = n AND spore-print-color = w) OR (bruises = f AND odor = c) OR (odor = m AND spore-print-color = w) OR (spore-print-color = r) OR (odor = p AND stalk-shape = e) OR (odor = f) OR (gill-spacing = w AND population = c)
\item (gill-size = b AND spore-print-color = h) OR (bruises = f AND stalk-root = b AND stalk-color-below-ring = w) OR (stalk-surface-above-ring = k AND habitat = d) OR (bruises = f AND gill-color = b) OR (gill-spacing = w AND population = c) OR (gill-size = b AND spore-print-color = r) OR (bruises = t AND odor = p)
\item (odor = p AND gill-attachment = f) OR (odor = c AND stalk-surface-above-ring = s) OR (gill-color = b) OR (odor = f) OR (gill-spacing = w AND population = c) OR (stalk-shape = e AND stalk-surface-above-ring = k AND stalk-surface-below-ring = y AND veil-color = w) OR (spore-print-color = r)
\end{enumerate}

\bibliography{bibliografia}

\begin{thebibliography}{}

\bibitem[Carlin and Chib, 1995]{Carlin1995}
Carlin, B.~P. and Chib, S. (1995).
\newblock Bayesian model choice via markov chain monte carlo methods.
\newblock {\em Journal of the Royal Statistical Society. Series B
  (Methodological)}, 57(3):473--484.

\bibitem[Consonni and Veronese, 1995]{Consonni1995}
Consonni, G. and Veronese, P. (1995).
\newblock A bayesian method for combining results from several binomial
  experiments.
\newblock {\em Journal of the American Statistical Association},
  90(431):935--944.

\bibitem[Green, 1995]{Green1995}
Green, P.~J. (1995).
\newblock Reversible jump markov chain monte carlo computation and bayesian
  model determination.
\newblock {\em Biometrika}, 82(4):711--732.

\bibitem[Hastie and Green, 2012]{Hastie2012}
Hastie, D.~I. and Green, P.~J. (2012).
\newblock Model choice using reversible jump markov chain monte carlo.
\newblock {\em Statistica Neerlandica}, 66(3):309--338.

\bibitem[Haws et~al., 2015]{Haws2015}
Haws, D.~C., Rish, I., Teyssedre, S., He, D., Lozano, A.~C., Kambadur, P.,
  Karaman, Z., and Parida, L. (2015).
\newblock Variable-selection emerges on top in empirical comparison of
  whole-genome complex-trait prediction methods.
\newblock {\em PLoS ONE}, 10.

\bibitem[Ishwaran and Rao, 2005]{Ishwaran2005}
Ishwaran, H. and Rao, J.~S. (2005).
\newblock Spike and slab variable selection: Frequentist and bayesian
  strategies.
\newblock {\em Ann. Statist.}, 33(2):730--773.

\bibitem[Kirkpatrick et~al., 1983]{Kirkpatrick1983}
Kirkpatrick, S., Gelatt, C.~D., and Vecchi, M.~P. (1983).
\newblock Optimization by simulated annealing.
\newblock {\em Science}, 220(4598):671--680.

\bibitem[Mavaddat et~al., 2019]{Mavaddat2019}
Mavaddat, N., Michailidou, K., Dennis, J., Lush, M., Fachal, L., Lee, A.,
  Tyrer, J.~P., Chen, T.-H., Wang, Q., Bolla, M.~K., Yang, X., Adank, M.~A.,
  Ahearn, T., Aittomäki, K., Allen, J., Andrulis, I.~L., Anton-Culver, H.,
  Antonenkova, N.~N., Arndt, V., Aronson, K.~J., Auer, P.~L., Auvinen, P.,
  Barrdahl, M., Freeman, L. E.~B., Beckmann, M.~W., Behrens, S., Benitez, J.,
  Bermisheva, M., Bernstein, L., Blomqvist, C., Bogdanova, N.~V., Bojesen,
  S.~E., Bonanni, B., Børresen-Dale, A.-L., Brauch, H., Bremer, M., Brenner,
  H., Brentnall, A., Brock, I.~W., Brooks-Wilson, A., Brucker, S.~Y., Brüning,
  T., Burwinkel, B., Campa, D., Carter, B.~D., Castelao, J.~E., Chanock, S.~J.,
  Chlebowski, R., Christiansen, H., Clarke, C.~L., Collée, J.~M.,
  Cordina-Duverger, E., Cornelissen, S., Couch, F.~J., Cox, A., Cross, S.~S.,
  Czene, K., Daly, M.~B., Devilee, P., Dörk, T., dos Santos-Silva, I., Dumont,
  M., Durcan, L., Dwek, M., Eccles, D.~M., Ekici, A.~B., Eliassen, A.~H.,
  Ellberg, C., Engel, C., Eriksson, M., Evans, D.~G., Fasching, P.~A.,
  Figueroa, J., Fletcher, O., Flyger, H., Försti, A., Fritschi, L.,
  Gabrielson, M., Gago-Dominguez, M., Gapstur, S.~M., García-Sáenz, J.~A.,
  Gaudet, M.~M., Georgoulias, V., Giles, G.~G., Gilyazova, I.~R., Glendon, G.,
  Goldberg, M.~S., Goldgar, D.~E., González-Neira, A., Alnæs, G. I.~G., Grip,
  M., Gronwald, J., Grundy, A., Guénel, P., Haeberle, L., Hahnen, E., Haiman,
  C.~A., Håkansson, N., Hamann, U., Hankinson, S.~E., Harkness, E.~F., Hart,
  S.~N., He, W., Hein, A., Heyworth, J., Hillemanns, P., Hollestelle, A.,
  Hooning, M.~J., Hoover, R.~N., Hopper, J.~L., Howell, A., Huang, G.,
  Humphreys, K., Hunter, D.~J., Jakimovska, M., Jakubowska, A., Janni, W.,
  John, E.~M., Johnson, N., Jones, M.~E., Jukkola-Vuorinen, A., Jung, A.,
  Kaaks, R., Kaczmarek, K., Kataja, V., Keeman, R., Kerin, M.~J.,
  Khusnutdinova, E., Kiiski, J.~I., Knight, J.~A., Ko, Y.-D., Kosma, V.-M.,
  Koutros, S., Kristensen, V.~N., Krüger, U., Kühl, T., Lambrechts, D.,
  Marchand, L.~L., Lee, E., Lejbkowicz, F., Lilyquist, J., Lindblom, A.,
  Lindström, S., Lissowska, J., Lo, W.-Y., Loibl, S., Long, J., Lubiński, J.,
  Lux, M.~P., MacInnis, R.~J., Maishman, T., Makalic, E., Kostovska, I.~M.,
  Mannermaa, A., Manoukian, S., Margolin, S., Martens, J.~W., Martinez, M.~E.,
  Mavroudis, D., McLean, C., Meindl, A., Menon, U., Middha, P., Miller, N.,
  Moreno, F., Mulligan, A.~M., Mulot, C., Muñoz-Garzon, V.~M., Neuhausen,
  S.~L., Nevanlinna, H., Neven, P., Newman, W.~G., Nielsen, S.~F.,
  Nordestgaard, B.~G., Norman, A., Offit, K., Olson, J.~E., Olsson, H., Orr,
  N., Pankratz, V.~S., Park-Simon, T.-W., Perez, J.~I., Pérez-Barrios, C.,
  Peterlongo, P., Peto, J., Pinchev, M., Plaseska-Karanfilska, D., Polley,
  E.~C., Prentice, R., Presneau, N., Prokofyeva, D., Purrington, K., Pylkäs,
  K., Rack, B., Radice, P., Rau-Murthy, R., Rennert, G., Rennert, H.~S.,
  Rhenius, V., Robson, M., Romero, A., Ruddy, K.~J., Ruebner, M., Saloustros,
  E., Sandler, D.~P., Sawyer, E.~J., Schmidt, D.~F., Schmutzler, R.~K.,
  Schneeweiss, A., Schoemaker, M.~J., Schumacher, F., Schürmann, P.,
  Schwentner, L., Scott, C., Scott, R.~J., Seynaeve, C., Shah, M., Sherman,
  M.~E., Shrubsole, M.~J., Shu, X.-O., Slager, S., Smeets, A., Sohn, C., Soucy,
  P., Southey, M.~C., Spinelli, J.~J., Stegmaier, C., Stone, J., Swerdlow,
  A.~J., Tamimi, R.~M., Tapper, W.~J., Taylor, J.~A., Terry, M.~B., Thöne, K.,
  Tollenaar, R.~A., Tomlinson, I., Truong, T., Tzardi, M., Ulmer, H.-U., Untch,
  M., Vachon, C.~M., van Veen, E.~M., Vijai, J., Weinberg, C.~R., Wendt, C.,
  Whittemore, A.~S., Wildiers, H., Willett, W., Winqvist, R., Wolk, A., Yang,
  X.~R., Yannoukakos, D., Zhang, Y., Zheng, W., Ziogas, A., Dunning, A.~M.,
  Thompson, D.~J., Chenevix-Trench, G., Chang-Claude, J., Schmidt, M.~K., Hall,
  P., Milne, R.~L., Pharoah, P.~D., Antoniou, A.~C., Chatterjee, N., Kraft, P.,
  García-Closas, M., Simard, J., and Easton, D.~F. (2019).
\newblock Polygenic risk scores for prediction of breast cancer and breast
  cancer subtypes.
\newblock {\em The American Journal of Human Genetics}, 104(1):21--34.

\bibitem[Muselli and Quarati, 2005]{Muselli2005}
Muselli, M. and Quarati, A. (2005).
\newblock Reconstructing positive boolean functions with shadow clustering.
\newblock In {\em Proceedings of the 2005 European Conference on Circuit Theory
  and Design, 2005.}, volume~3, pages III/377--III/380 vol. 3.

\bibitem[Roberts and Rosenthal, 2016]{Roberts2016}
Roberts, G.~O. and Rosenthal, J.~S. (2016).
\newblock Complexity bounds for markov chain monte carlo algorithms via
  diffusion limits.
\newblock {\em Journal of Applied Probability}, 53(2):410--420.

\bibitem[{Shang} et~al., 2019]{Shang2019}
{Shang}, J., {Wang}, X., {Wu}, X., {Sun}, Y., {Ding}, Q., {Liu}, J., and
  {Zhang}, H. (2019).
\newblock A review of ant colony optimization based methods for detecting
  epistatic interactions.
\newblock {\em IEEE Access}, 7:13497--13509.

\bibitem[Sisson, 2005]{Sisson2005}
Sisson, S.~A. (2005).
\newblock Transdimensional markov chains: A decade of progress and future
  perspectives.
\newblock {\em Journal of the American Statistical Association},
  100(471):1077--1089.

\bibitem[Stephens, 2000]{Stephens2000}
Stephens, M. (2000).
\newblock Bayesian analysis of mixture models with an unknown number of
  components—an alternative to reversible jump methods.
\newblock {\em Ann. Statist.}, 28(1):40--74.

\bibitem[Yin et~al., 2019]{Yin2019}
Yin, B., Balvert, M., van~der Spek, R. A.~A., Dutilh, B.~E., Bohté, S.,
  Veldink, J., and Schönhuth, A. (2019).
\newblock {Using the structure of genome data in the design of deep neural
  networks for predicting amyotrophic lateral sclerosis from genotype}.
\newblock {\em Bioinformatics}, 35(14):i538--i547.

\end{thebibliography}
\bibliographystyle{apalike}

\end{document}